%% file: acl2021.tex
\documentclass[11pt,a4paper]{article}
\usepackage[hyperref]{acl2021}
\usepackage{times}
\usepackage{latexsym}

\usepackage{graphicx}
\usepackage{microtype}
\usepackage{amsmath}
\usepackage{amsfonts}
\usepackage{booktabs}
\usepackage{threeparttable}
\usepackage{mathrsfs}
\usepackage{latexsym}
\usepackage{stmaryrd}
\usepackage{multicol}
\usepackage{multirow}
\usepackage{cleveref}
\usepackage{xcolor}

\crefname{section}{§}{§§}
\Crefname{section}{§}{§§}
\aclfinalcopy 
\def\aclpaperid{3798} 

\title{LGESQL: Line Graph Enhanced Text-to-SQL Model with Mixed Local and Non-Local Relations}

\author{Ruisheng Cao$^{1}$, Lu Chen$^{1,2*}$, Zhi Chen$^{1}$, Yanbin Zhao$^{1}$,\\
\textbf{Su Zhu$^{3}$ and Kai Yu$^{1,2}$}\thanks{\ \ The corresponding authors are Lu Chen and Kai Yu.}\\
  $^{1}$X-LANCE Lab, Department of Computer Science and Engineering\\
  MoE Key Lab of Artificial Intelligence, AI Institute, Shanghai Jiao Tong University\\
  Shanghai Jiao Tong University, Shanghai, China\\
  $^{2}$State Key Lab of Media Convergence Production Technology and Systems, Beijing, China\\
  $^{3}$AISpeech Co., Ltd., Suzhou, China\\
  {\tt \{211314,chenlusz,kai.yu\}@sjtu.edu.cn}\\}

\date{}

\begin{document}
\maketitle

\input{0.abstract}
\input{1.introduction}

\input{2.basic}
\input{3.model}
\input{4.experiment}
\input{5.related_work}
\input{6.conclusion}

\section*{Acknowledgments}
We thank Tao Yu, Yusen Zhang and Bo Pang for their careful assistance with the evaluation. We also thank the anonymous reviewers for their thoughtful comments. This work has been supported by Shanghai Municipal Science and Technology Major Project~(2021SHZDZX0102),  No.SKLMCPTS2020003 Project and Startup Fund for Youngman Research at SJTU (SFYR at SJTU).

\bibliographystyle{acl_natbib}
\bibliography{acl2021}
\clearpage
\appendix
\input{7.appendix}

\end{document}

%% file: 0.abstract.tex
\begin{abstract}
This work aims to tackle the challenging heterogeneous graph encoding problem in the text-to-SQL task. Previous methods are typically node-centric and merely utilize different weight matrices to parameterize edge types, which 1) ignore the rich semantics embedded in the topological structure of edges, and 2) fail to distinguish local and non-local relations for each node. To this end, we propose a \textbf{L}ine \textbf{G}raph \textbf{E}nhanced Text-to-\textbf{SQL}~(LGESQL) model to mine the underlying relational features without constructing meta-paths. By virtue of the line graph, messages propagate more efficiently through not only connections between nodes, but also the topology of directed edges. Furthermore, both local and non-local relations are integrated distinctively during the graph iteration. We also design an auxiliary task called \emph{graph pruning} to improve the discriminative capability of the encoder. Our framework achieves state-of-the-art results~($62.8\%$ with \textsc{Glove}, $72.0\%$ with \textsc{Electra}) on the cross-domain text-to-SQL benchmark Spider at the time of writing. 
\end{abstract}

%% file: 1.introduction.tex
\section{Introduction}
The text-to-SQL task~\cite{zhong2017seq2sql,xu2017sqlnet} aims to convert a natural language question into a SQL query, given the corresponding database schema. It has been widely studied in both academic and industrial communities to build natural language interfaces to databases~(NLIDB,~\citealp{androutsopoulos1995natural}).
\begin{figure}[t]
    \centering
    \includegraphics[width=0.48\textwidth]{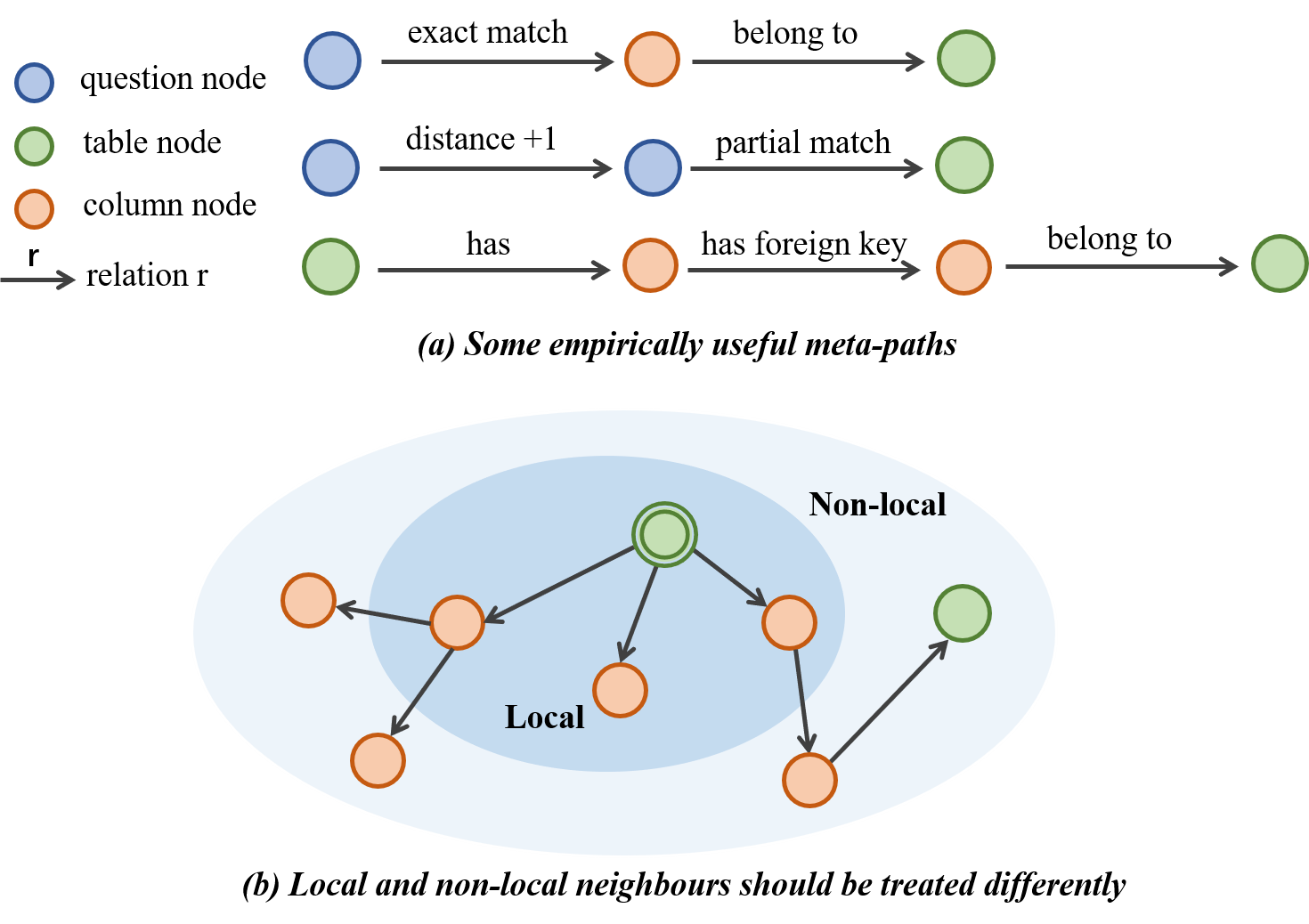}
    \caption{Two limitations if edge features are retrieved from a fixed-size embedding matrix: (a) fail to discover useful meta-paths, and (b) unable to differentiate local and non-local neighbors.}
    \label{fig:challenge}
\end{figure}

One daunting problem is how to jointly encode the question words and database schema items~(including tables and columns), as well as various relations among these heterogeneous inputs. Typically, previous literature utilizes a node-centric graph neural network~(GNN, \citealp{scarselli2008graph}) to aggregate information from neighboring nodes. GNNSQL~\cite{bogin-etal-2019-representing} adopts a relational graph convolution network~(RGCN,~\citealp{schlichtkrull2018modeling}) to take into account different edge types between schema items, such as \textsc{T-Has-C} relationship~\footnote{For abbreviation, Q represents \textsc{Question} node, while T and C represent \textsc{Table} and \textsc{Column} nodes.}, primary key and foreign key constraints. However, these edge features are directly retrieved from a fixed-size parameter matrix and may suffer from the drawback: unaware of contextualized information, especially the structural topology of edges. Meta-path is defined as a composite relation linking two objects, which can be used to capture multi-hop semantics. For example, in Figure \ref{fig:challenge}(a), relation \textsc{Q-ExactMatch-C} and \textsc{C-BelongsTo-T} can form a 2-hop meta-path indicating that some table $t$ has one column exactly mentioned in the question.

Although RATSQL~\cite{wang-etal-2020-rat} introduces some useful meta-paths such as \textsc{C-SameTable-C}, it treats all relations, either $1$-hop or multi-hop, in the same manner~(relative position embedding, \citealp{shaw-etal-2018-self}) in a complete graph. Without distinguishing local and non-local neighbors, see Figure \ref{fig:challenge}(b), each node will attend to all the other nodes equally, which may lead to the notorious over-smoothing problem~\cite{chen2020measuring}. Besides, meta-paths are currently constructed by domain experts or explored by breadth-first search~\cite{kong2012meta}. Unfortunately, the number of possible meta-paths increases exponentially with the path length, and selecting the most important subset among them is an NP-complete problem~\cite{lao2010relational}.

To address the above limitations, we propose a \textbf{L}ine \textbf{G}raph \textbf{E}nhanced Text-to-\textbf{SQL} model~(LGESQL), which explicitly considers the topological structure of edges. According to the definition of a line graph~\cite{gross2005graph}, we firstly construct an edge-centric graph from the original node-centric graph. These two graphs capture the structural topology of nodes and edges, respectively. Iteratively, each node in either graph gathers information from its neighborhood and incorporates edge features from the dual graph to update its representation. As for the node-centric graph, we combine both local and non-local edge features into the computation. Local edge features denote $1$-hop relations and are dynamically provided by node embeddings in the line graph, while non-local edge features are directly extracted from a parameter matrix. This distinction encourages the model to pay more attention to local edge features while maintaining information from multi-hop neighbors. Additionally, we propose an auxiliary task called \emph{graph pruning}. It introduces an inductive bias that the heterogeneous graph encoder of text-to-SQL should be intelligent to extract the golden schema items related to the question from the entire database schema graph.

Experimental results on benchmark Spider~\cite{yu-etal-2018-spider} demonstrate that our LGESQL model promotes the exact set match accuracy to $62.8\%$~(with \textsc{GloVe},~\citeauthor{pennington-etal-2014-glove}~\citeyear{pennington-etal-2014-glove}) and $72.0\%$~(with pretrained language model \textsc{Electra},~\citeauthor{DBLP:conf/iclr/ClarkLLM20}~\citeyear{DBLP:conf/iclr/ClarkLLM20}). Our main contributions are summarized as follows:
\begin{itemize}
    \item We propose to model the $1$-hop edge features with a line graph in text-to-SQL. Both non-local and local features are integrated during the iteration process of node embeddings.
    \item We design an auxiliary task called \emph{graph pruning}, which aims to determine whether each node in the database schema graph is relevant to the given question.
    \item Empirical results on dataset Spider demonstrate that our model is effective, and we achieve state-of-the-art performances both without and with pre-trained language models.
\end{itemize}

%% file: 2.basic.tex
\section{Preliminaries}
\paragraph{Problem definition}
Given a natural language question $Q=(q_1,q_2,\cdots,q_{|Q|})$ with length $|Q|$ and the corresponding database schema $S=T\cup C$, the target is to generate a SQL query $y$. The database schema $S$ contains multiple tables $T=\{t_1,t_2,\cdots\}$ and columns $C=\{c_1^{t_1},c_2^{t_1},\cdots,c_1^{t_{2}},c_2^{t_{2}},\cdots\}$. Each table $t_{i}$ is described by its name and is further composed of several words $(t_{i1}, t_{i2},\cdots)$. Similarly, we use word phrase $(c_{j1}^{t_i},c_{j2}^{t_i},\cdots)$ to represent column $c_j^{t_i}\in t_i$. Besides, each column $c_j^{t_i}$ also has a type field $c_{j0}^{t_i}$ to constrain its cell values~(e.g. \textsc{Text} and \textsc{Number}). 

The entire input node-centric heterogeneous graph $G^n=(V^n, R^n)$ consists of all three types of nodes mentioned above, that is $V^n=Q\cup T\cup C$ with the number of nodes $|V^n|=|Q|+|T|+|C|$, where $|T|$ and $|C|$ are the number of tables and columns respectively.

\paragraph{Meta-path} As shown in Figure \ref{fig:challenge}(a), a meta-path represents a path $\tau_1\overset{r_1}{\rightarrow}\tau_2\overset{r_2}{\rightarrow}\cdots\overset{r_l}{\rightarrow}\tau_{l+1}$, where the target vertex type of previous relation $r_{i-1}$ equals to the source vertex type $\tau_i$ of the current relation $r_i$. It describes a composite relation $r=r_1\circ r_2\cdots \circ r_l$ between nodes with type $\tau_1$ and $\tau_{l+1}$. In this work, $\tau_i\in \{\textsc{Question,Table,Column}\}$. Throughout our discussion, we use the term \emph{local} to denote relations with path length $1$, while \emph{non-local} relations refer to meta-paths longer than $1$. The relational adjacency matrix $R^n$ contains both local and non-local relations, see Appendix \ref{app:relations} for enumeration.

\begin{figure}[htbp]
    \centering
    \includegraphics[width=0.48\textwidth]{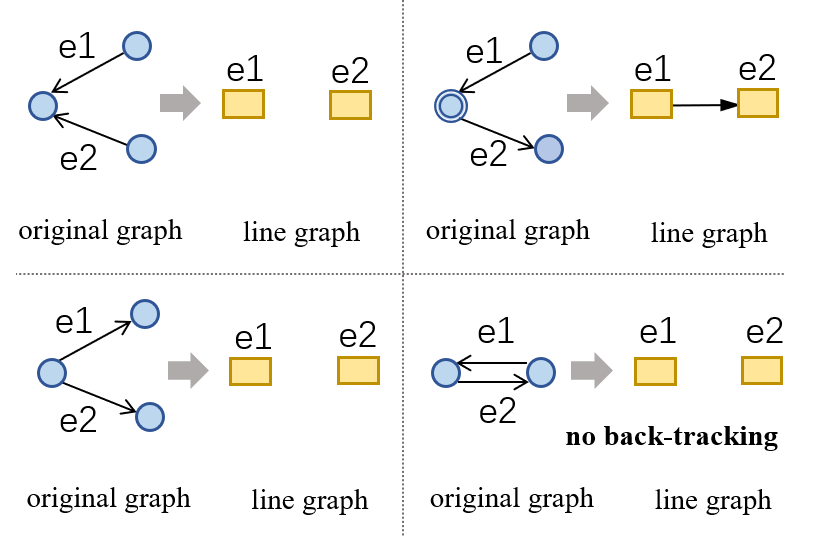}
    \caption{Construction of a line graph. For clarity, we simplify the notation of edges.}
    \label{fig:line_graph}
\end{figure}

\paragraph{Line Graph} Each vertex $v_i^{e},i=1,2,\cdots,|V^e|$ in the line graph $G^{e}=(V^{e},R^e)$ can be uniquely mapped to a directed edge $r_{st}^n\in R^n$, or $v_s^n\rightarrow v_t^n$, in the original node-centric graph $G^n=(V^n, R^n)$. Function $f$ maps the source and target node index tuple $(s,t)$ into the ``edge" index $i=f(s,t)$ in $G^e$. The reverse mapping is $f^{\textrm{-}1}$. In the line graph $G^e$, a directed edge $r^e_{ij}\in R^e$ exists from node $v^e_i$ to $v^e_j$, iff the target node of edge $r^n_{f^{\textrm{-}1}(i)}$ and the source node of edge $r^n_{f^{\textrm{-}1}(j)}$ in $G^n$ are exactly the same node. Actually, $r^e_{ij}$ captures the information flow in meta-path $r^n_{f^{\textrm{-}1}(i)}\circ r^n_{f^{\textrm{-}1}(j)}$. We prevent back-tracking cases where two reverse edges will not be connected in $G^e$, illustrated in Figure \ref{fig:line_graph}.

We only utilize local relations in $R^n$ as the node set $V^e$ to avoid creating too many nodes in the line graph $G^e$. Symmetrically, each edge in $R^e$ can be uniquely identified by the node in $V^n$. For example, in the upper right part of Figure \ref{fig:line_graph}, the edge between nodes ``e1" and ``e2" in the line graph can be represented by the middle node with double solid borderlines in the original graph.

%% file: 3.model.tex
\section{Method}
\label{sec:method}
\begin{figure*}[htbp]
    \centering
    \includegraphics[width=0.95\textwidth]{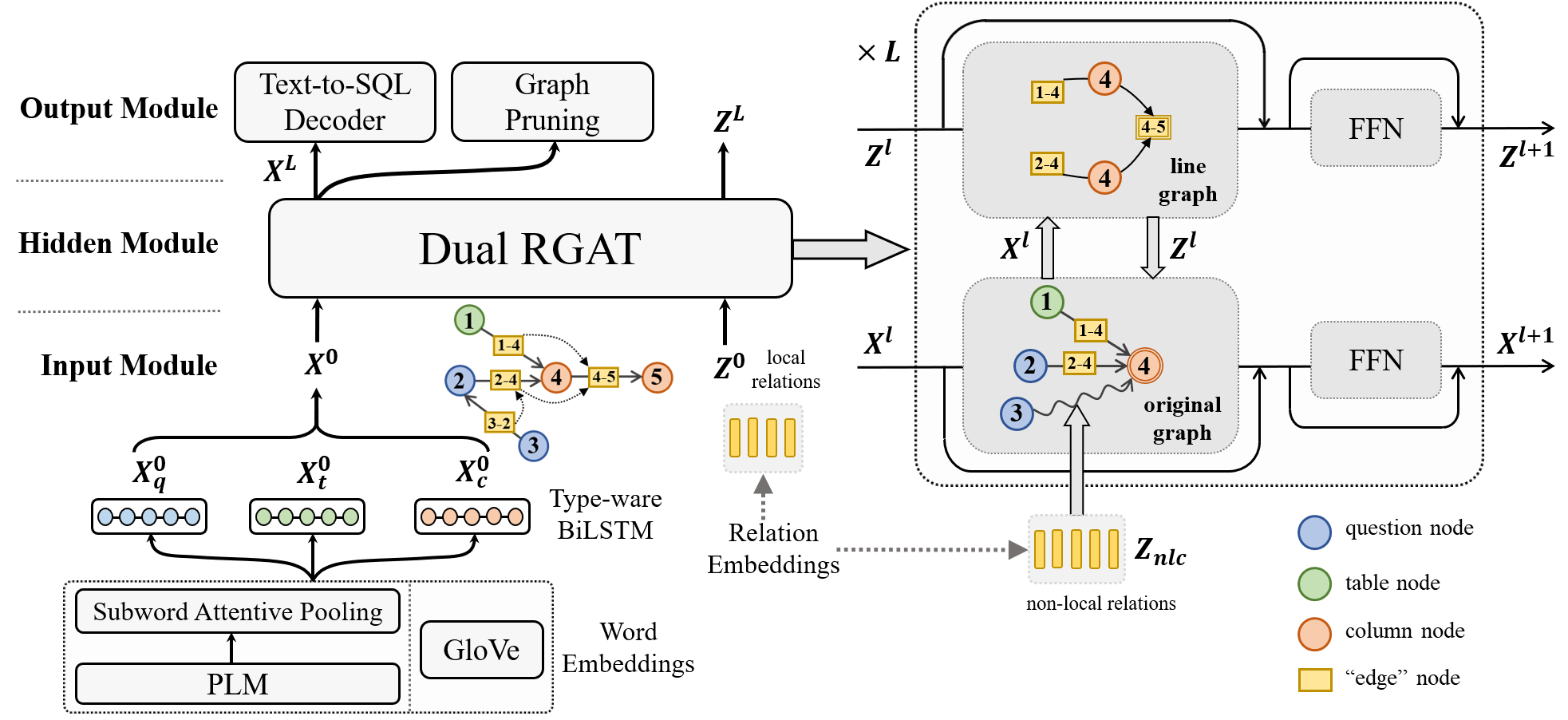}
    \caption{The overall model architecture. We use bidirectional edges in practice but only draw unidirectional edges for better understanding. In the Dual RGAT module, we take the node with index $4$ and the edge with label $4\textrm{-}5$ as the main focuses.}
    \label{fig:model}
\end{figure*}

After constructing the line graph, we utilize the classic encoder-decoder architecture~\cite{DBLP:conf/nips/SutskeverVL14,DBLP:journals/corr/BahdanauCB14} as the backbone of our model. LGESQL consists of three parts: a graph input module, a line graph enhanced hidden module, and a graph output module~(see Figure \ref{fig:model} for an overview). The first two modules aim to map the input heterogeneous graph $G^n$ into node embeddings $\mathbf{X}\in\mathbb{R}^{|V^n|\times d}$, where $d$ is the graph hidden size. The graph output module retrieves and transforms $\mathbf{X}$ into the target SQL query $y$.

\subsection{Graph Input Module}
This module aims to provide the initial embeddings for both nodes and edges. Initial local edge features $\mathbf{Z}^0\in\mathbb{R}^{|V^e|\times d}$ and non-local edge features $\mathbf{Z}_{nlc}\in\mathbb{R}^{(|R^n|-|V^e|)\times d}$ are directly retrieved from a parameter matrix. For nodes, we can obtain their representations from either word vectors  \textsc{Glove}~\cite{pennington-etal-2014-glove} or a pre-trained language model~(PLM) such as \textsc{Bert}~\cite{devlin-etal-2019-bert}.
\paragraph{\textsc{GloVe}}
Each word $q_i$ in the question $Q$ or schema item $t_i\in T$ or $c_j^{t_i}\in C$ can be initialized by looking up the embedding dictionary without considering the context. Then, these vectors are passed into three type-ware bidirectional LSTMs~(BiLSTM,~\citealp{hochreiter1997long}) respectively to attain contextual information. We concatenate the forward and backward hidden states for each question word $q_i$ as the graph input $\mathbf{x}^0_{q_i}$. As for table $t_i$, after feeding $(t_{i0}, t_{i1}, t_{i2}, \cdots)$ into the BiLSTM~(special type $t_{i0}=``table",\forall i$), we concatenate the last hidden states in both directions as the graph input $\mathbf{x}^0_{t_i}$~(similarly for column $c_j^{t_i}$). These node representations are stacked together to form the initial node embeddings matrix $\mathbf{X}^0\in\mathbb{R}^{|V^n|\times d}$.

\paragraph{PLM} Firstly, we flatten all question words and schema items into a sequence, where columns belong to the same table are clustered together~\footnote{Following \citet{suhr-etal-2020-exploring}, we randomly shuffle the order of tables and columns in different mini-batches to discourage over-fitting.}: {\tt [CLS]$q_1q_2\cdots q_{|Q|}$[SEP]$t_{10}t_1 c_{10}^{t_1}c_{1}^{t_1}c_{20}^{t_1}c_{2}^{t_1}\cdots$ $t_{20}t_2c_{10}^{t_2}c_{1}^{t_2}c_{20}^{t_2}c_{2}^{t_2}\cdots$[SEP]}. The type information $t_{i0}$ or $c_{j0}^{t_i}$ is inserted before each schema item. Since each word $w$ is tokenized into sub-words, we append a subword attentive pooling layer after PLM to obtain word-level representations. Concretely, given the output sequence of subword features $\mathbf{w}^s_1,\mathbf{w}^s_2,\cdots,\mathbf{w}^s_{|w|}$ for each subword $w^s_i$ in $w$, the word-level representation $\mathbf{w}$ is~\footnote{Vectors throughout this paper are all row vectors.}
\begin{align*}
a_i=&\text{softmax}_i\ \text{tanh}(\mathbf{w}^s_i\mathbf{W}_s)\mathbf{v}_s^{\mathrm{T}},\\
\mathbf{w}=&\sum_{i}a_i\mathbf{w}^s_i,
\end{align*}
where $\mathbf{v}_s$ and $\mathbf{W}_s$ are trainable parameters. After obtaining the word vectors, we also feed them into three BiLSTMs according to the node types and get the graph inputs $\mathbf{X}^0$ for all nodes.

\subsection{Line Graph Enhanced Hidden Module}
It contains a stack of $L$ dual relational graph attention network~(Dual RGAT) layers. In each layer $l$, two RGATs~\cite{wang-etal-2020-relational} capture the structure of the original graph and line graph, respectively. Node embeddings in one graph play the role of edge features in another graph. For example, the edge features used in graph $G^n$ are provided by the node embeddings in graph $G^e$.

We use $\mathbf{X}^l\in \mathbb{R}^{|V^n|\times d}$ to denote the input node embedding matrix of graph $G^n$ in the $l$-th layer, $l\in\{0,1,\cdots,L-1\}$. As for each specific node $v^n_i\in V^n$, we use $\mathbf{x}^l_i$. Similarly, matrix $\mathbf{Z}^l\in\mathbb{R}^{|V^e|\times d}$ and vector $\mathbf{z}^l_i$ are used to denote node embeddings in the line graph. Following RATSQL~\cite{wang-etal-2020-rat}, we use multi-head scaled dot-product~\cite{vaswani2017attention} to calculate the attention weights. For brevity, we formulate the entire computation in one layer as two basic modules:
\begin{align*}
\mathbf{X}^{l+1}=&\text{RGAT}^n(\mathbf{X}^l,[\mathbf{Z}^{l};\mathbf{Z}_{nlc}], G^n),\\
\mathbf{Z}^{l+1}=&\text{RGAT}^e(\mathbf{Z}^l,\mathbf{X}^{l}, G^e),
\end{align*}
where $\mathbf{Z}_{nlc}$ is the aforementioned non-local edge features in the original graph $G^n$.

\subsubsection{RGAT for the Original Graph}
\label{sec:mix}
\begin{figure}[htbp]
    \centering
    \includegraphics[width=0.40\textwidth]{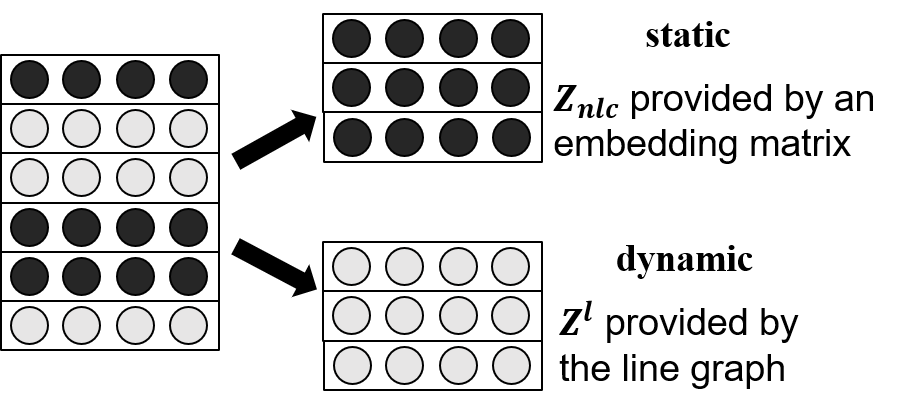}
    \caption{Mixed static and dynamic embeddings.}
    \label{fig:msde}
\end{figure}

Given the node-centric graph $G^n$, the output representation $\mathbf{x}^{l+1}_i$  of the $l$-th layer is computed by
\begin{align*}
\tilde{\alpha}_{ji}^h=&(\mathbf{x}^{l}_i\mathbf{W}^h_q)(\mathbf{x}^{l}_j\mathbf{W}^h_k + [\psi(r_{ji}^n)]^H_h)^{\mathrm{T}},\\
\alpha_{ji}^h=&\text{softmax}_j(\tilde{\alpha}_{ji}^h/\sqrt{d/H}),\\
\tilde{\mathbf{x}}^{l}_i=&\bigparallel_{h=1}^H\sum_{v_j^n\in\mathcal{N}^n_i}\alpha_{ji}^h(\mathbf{x}^{l}_j\mathbf{W}^h_v + [\psi(r_{ji}^n)]^H_h),\\
\tilde{\mathbf{x}}^{l+1}_i=&\text{LayerNorm}(\mathbf{x}^{l}_i+\tilde{\mathbf{x}}^{l}_i\mathbf{W}_o),\\
\mathbf{x}^{l+1}_i=&\text{LayerNorm}(\tilde{\mathbf{x}}^{l+1}_i+\text{FFN}(\tilde{\mathbf{x}}^{l+1}_i)),
\end{align*}
where $\parallel$ represents vector concatenation, matrices $\mathbf{W}_{q}^h,\mathbf{W}_{k}^h,\mathbf{W}_{v}^h\in \mathbb{R}^{d\times d/H},\mathbf{W}_{o}\in \mathbb{R}^{d\times d}$ are trainable parameters, $H$ is the number of heads and $\text{FFN}(\cdot)$ denotes a feedforward neural network. $\mathcal{N}^n_i$ represents the receptive field of node $v^n_i$ and function $\psi(r_{ji}^n)$ returns a $d$-dim feature vector of relation $r_{ji}^n$. Operator $[\cdot]_h^H$ first evenly splits the vector into $H$ parts and returns the $h$-th partition. Since there are two genres of relations~(local and non-local), we design two schemes to integrate them:
\paragraph{Mixed Static and Dynamic Embeddings}
If $r_{ji}^n$ is a local relation, $\psi(r_{ji}^n)$ returns the node embedding $\mathbf{z}^l_{f(j,i)}$ from the line graph\footnote{Function $f$ maps the tuple of source and target node indices in $G^n$ into the corresponding node index in $G^e$.}. Otherwise, $\psi(r_{ji}^n)$ directly retrieves the vector from the non-local embedding matrix $\mathbf{Z}_{nlc}$, see Figure \ref{fig:msde}. The neighborhood function $\mathcal{N}^n_i$ for node $v^n_i$ returns the entire node set $V^n$ and is shared across different heads.

\begin{figure}[htbp]
    \centering
    \includegraphics[width=0.40\textwidth]{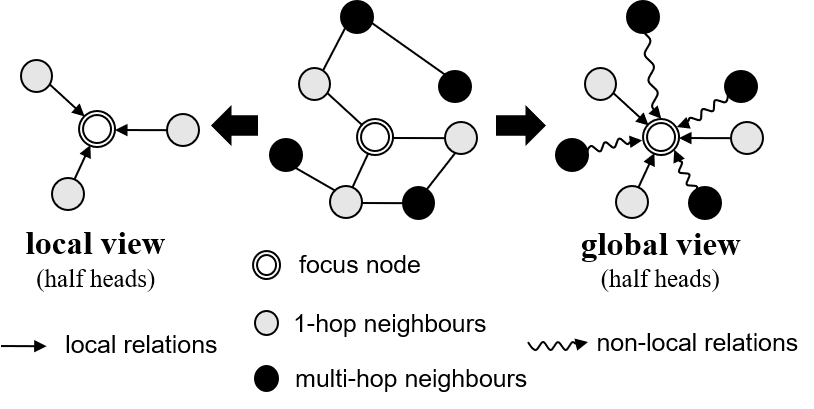}
    \caption{Multi-head multi-view concatenation.}
    \label{fig:mmc}
\end{figure}

\paragraph{Multi-head Multi-view Concatenation}
An alternative is to split the muli-head attention module into two parts. In half of the heads, the neighborhood function $\mathcal{N}^n_i$ of node $v^n_i$ only contains nodes that are reachable within $1$-hop. In this case, $\psi(r_{ji}^n)$ returns the layer-wise updated feature $\mathbf{z}^l_{f(j,i)}$ from $\mathbf{Z}^l$. In the other heads, each node has access to both local and non-local neighbors, and $\psi(\cdot)$ always returns static entries in the embedding matrix $\mathbf{Z}_{nlc}\cup\mathbf{Z}^0$, see Figure \ref{fig:mmc} for illustration.

In either scheme, the RGAT module treats local and non-local relations differently and relatively manipulates the local edge features more carefully.

\subsubsection{RGAT for the Line Graph}
Symmetrically, given edge-centric graph $G^e$, the updated node representation $\mathbf{z}^{l+1}_i$ from $\mathbf{z}^l_i$ is calculated similarly with little modifications:
\begin{align*}
\tilde{\beta}_{ji}^h=&(\mathbf{z}^l_i\mathbf{U}^h_q + [\phi(r_{ji}^e)]^H_h)(\mathbf{z}^l_j\mathbf{U}^h_k)^{\mathrm{T}},\\
\beta_{ji}^h=&\text{softmax}_j(\tilde{\beta}_{ji}^h/\sqrt{d/H}),\\
\tilde{\mathbf{z}}^l_i=&\bigparallel_{h=1}^H\sum_{v_j^e\in\mathcal{N}^e_i}\beta_{ji}^h(\mathbf{z}^l_j\mathbf{U}^h_v + [\phi(r_{ji}^e)]^H_h),\\
\tilde{\mathbf{z}}^{l+1}_i=&\text{LayerNorm}(\mathbf{z}^l_i+\tilde{\mathbf{z}}^l_i\mathbf{U}_o),\\
\mathbf{z}^{l+1}_i=&\text{LayerNorm}(\tilde{\mathbf{z}}^{l+1}_i+\text{FFN}(\tilde{\mathbf{z}}^{l+1}_i)).
\end{align*}
Here $\phi(r_{ji}^e)$ returns the feature vector of relation $r_{ji}^e$ in $G^e$. Since we only consider local relations in the line graph, $\mathcal{N}^e_i$ only includes $1$-hop neighbous and $\phi(r_{ji}^e)$ equals to the source node embedding in $\mathbf{X}^l$ of edge $v_i^e$. Attention that the relational feature is added on the ``query" side instead of the ``key" side when computing attention logits $\tilde{\beta}_{ji}^h$ cause it is irrelevant to the incoming edges. For example, in Figure \ref{fig:model}, the connecting nodes of two edge pairs $(1\textrm{-}4,4\textrm{-}5)$ and $(2\textrm{-}4,4\textrm{-}5)$ are the same node with index $4$. $\mathbf{U}_{q}^h,\mathbf{U}_{k}^h,\mathbf{U}_{v}^h\in \mathbb{R}^{d\times d/H},\mathbf{U}_{o}\in \mathbb{R}^{d\times d}$ are trainable parameters.

The output matrices of the final layer $L$ are the desired outputs of the encoder: $\mathbf{X}=\mathbf{X}^L,\mathbf{Z}=\mathbf{Z}^L$.

\subsection{Graph Output Module}
This module includes two tasks: one decoder for the main focus text-to-SQL and the other one to perform an auxiliary task called \emph{graph pruning}. We use the subscript to denote the collection of node embeddings with a specific type, e.g., $\mathbf{X}_q$ is the matrix of all question node embeddings.
\subsubsection{Text-to-SQL Decoder}
We adopt the grammar-based syntactic neural decoder~\cite{yin-neubig-2017-syntactic} to generate the abstract syntax tree~(AST) of the target query $y$ in depth-first-search order. The output at each decoding timestep is either 1) an \textsc{ApplyRule} action that expands the current non-terminal node in the partially generated AST, or 2) \textsc{SelectTable} or \textsc{SelectColumn} action that chooses one schema item $\mathbf{x}_{s_i}$ from the encoded memory $\mathbf{X}_s=\mathbf{X}_t\cup\mathbf{X}_c$. Mathematically, $P(y|\mathbf{X})=\prod_{j}P(a_j|a_{<j},\mathbf{X})$, where $a_j$ is the action at the $j$-th timestep. For more implementation details, see Appendix \ref{app:decoder}.

\subsubsection{Graph Pruning}
\begin{figure}[htbp]
    \centering
    \includegraphics[width=0.40\textwidth]{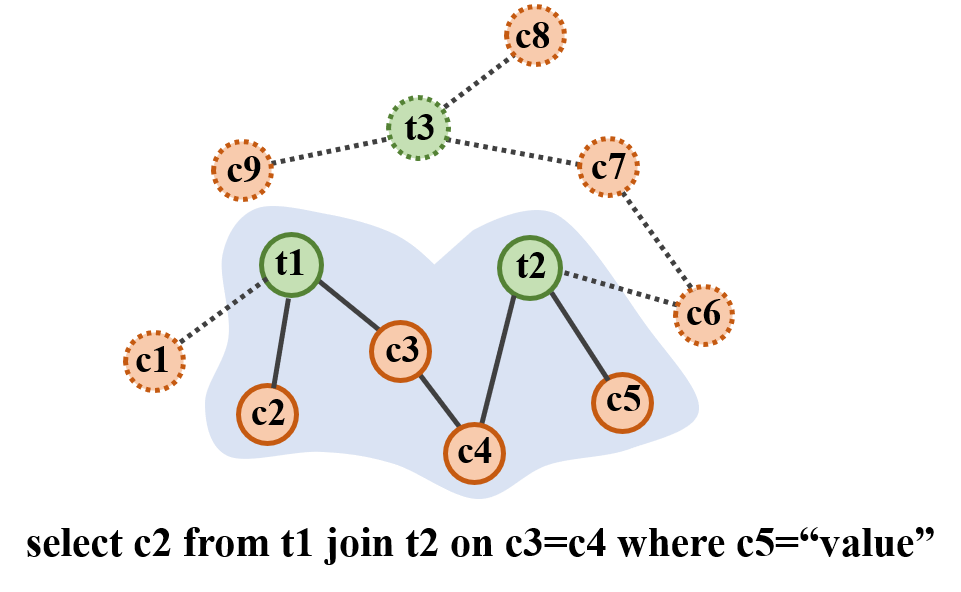}
    \caption{A delexicalized example of graph pruning. Circles with dashed borderlines are irrelevant schema items, thus labeled with $0$.}
    \label{fig:graph_pruning}
\end{figure}

We hypothesize that a powerful encoder should distinguish irrelevant schema items from golden schema items used in the target query. In Figure \ref{fig:graph_pruning}, the question-oriented schema sub-graph~(above the shadow region) can be easily extracted. The intent $c2$ and the constraint $c5$ are usually explicitly mentioned in the question, identified by dot-product attention mechanism or schema linking. The linking nodes such as $t1,c3,c4,t2$ can be inferred by the $1$-hop connections of the schema graph to form a connected component. To introduce this inductive bias, we design an auxiliary task that aims to classify each schema node $s_i\in S=T\cup C$ based on its relevance with the question and the sparse structure of the schema graph.

Firstly, we compute the context vector $\tilde{\mathbf{x}}_{s_i}$ from the question node embeddings $\mathbf{X}_q$ for each schema node $s_i$ via multi-head attention.
\begin{align*}
\gamma^h_{ji}=&\text{softmax}_j\frac{(\mathbf{x}_{s_i}\mathbf{W}_{sq}^h)(\mathbf{x}_{q_j}\mathbf{W}_{sk}^h)^{\mathrm{T}}}{\sqrt{d/H}},\\
\tilde{\mathbf{x}}_{s_i}=&(\bigparallel_{h=1}^H\sum_{j}\gamma^h_{ji} \mathbf{x}_{q_j}\mathbf{W}^h_{sv})\mathbf{W}_{so},
\end{align*}
where $\mathbf{W}_{sq}^h,\mathbf{W}_{sk}^h,\mathbf{W}^h_{sv}\in\mathbb{R}^{d\times d/H}$ and $\mathbf{W}_{so}\in\mathbb{R}^{d\times d}$ are network parameters. Then, a biaffine~\cite{DBLP:conf/iclr/DozatM17} binary classifier is used to determine whether the compressed context vector $\tilde{\mathbf{x}}_{s_i}$ and the schema node embedding $\mathbf{x}_{s_i}$ are correlated.
\begin{align*}
\text{Biaffine}(\mathbf{x}_1,\mathbf{x}_2)=&\mathbf{x}_1\mathbf{U}_{s}\mathbf{x}_2^\mathrm{T} + [\mathbf{x}_1;\mathbf{x}_2]\mathbf{W}_{s} + b_{s},\\
P^{\text{gp}}(y_{s_i}|\mathbf{x}_{s_i},\mathbf{X}_q)=&\sigma (\text{Biaffine}(\mathbf{x}_{s_i},\tilde{\mathbf{x}}_{s_i})).
\end{align*}
The ground truth label $y_{s_i}^g$ of a schema item is $1$ iff $s_i$ appears in the target SQL query. The training object can be formulated as
\begin{multline*}
\mathcal{L}_{gp}=-\sum_{s_i}[y_{s_i}^g\log P^{gp}(y_{s_i}|\mathbf{x}_{s_i},\mathbf{X}_q)\\
+(1-y_{s_i}^g)\log (1-P^{gp}(y_{s_i}|\mathbf{x}_{s_i},\mathbf{X}_q))].
\end{multline*}
This auxiliary task is combined with the main text-to-SQL task in a multitasking way. Similar ideas~\cite{bogin-etal-2019-global,DBLP:journals/corr/abs-2009-13845} and other association schemes are discussed in Appendix \ref{app:gp}.

%% file: 4.experiment.tex
\section{Experiments}
In this section, we evaluate our LGESQL model in different settings. Codes are public available~\footnote{\url{https://github.com/rhythmcao/text2sql-lgesql.git}.}.

\subsection{Experiment Setup}
\paragraph{Dataset} Spider~\cite{yu-etal-2018-spider} is a large-scale cross-domain zero-shot text-to-SQL benchmark~\footnote{Leaderboard of the challenge: \url{https://yale-lily.github.io//spider}.}. It contains $8659$ training examples across $146$ databases in total, and covers several domains from other datasets such as Restaurants~\cite{popescu2003towards}, GeoQuery~\cite{zelle1996learning}, Scholar~\cite{iyer-etal-2017-learning}, Academic~\cite{li2014constructing}, Yelp and IMDB~\cite{yaghmazadeh2017sqlizer} datasets. The detailed statistics are shown in Table \ref{tab:stats}. We follow the common practice to report the exact set match accuracy on the validation and test dataset. The test dataset contains $2147$ samples with $40$ unseen databases but is not public available. We submit our model to the organizer of the challenge for evaluation.
\begin{table}[htbp]
  \centering{\small
    \begin{tabular}{c|ccc}
    \toprule
      & \textbf{Train} & \textbf{Dev}\\
    \hline
    \# of samples & $8659$ & $1034$  \\
    \# of databases & $146$ & $20$ \\
    \hline
    Avg \# of question nodes & $13.4$ & $13.8$ \\
    Avg \# of table nodes & $6.6$ & $4.5$ \\
    Avg \# of column nodes & $33.1$ & $25.8$ \\
    Avg \# of nodes & $53.1$ & $44.1$ \\
    \hline
    Avg \# of actions & $16.3$ & $15.4$ \\
    \bottomrule
    \end{tabular}%
  \caption{Statistics for dataset Spider. The action sequence is created with our designed grammar.}
  \label{tab:stats}%
  }
\end{table}%
\paragraph{Implementations}
We preprocess the questions, table names, and column names with toolkit Stanza~\cite{qi-etal-2020-stanza} for tokenization and lemmatization. Our model is implemented with Pytorch~\cite{DBLP:conf/nips/PaszkeGMLBCKLGA19}, and the original and line graphs are constructed with library DGL~\cite{wang2019dgl}. Within the encoder, we use \textsc{GloVe}~\cite{pennington-etal-2014-glove} word embeddings with dimension $300$ or pretrained language models~(PLMs), \textsc{Bert}~\cite{devlin-etal-2019-bert} or \textsc{Electra}~\citep{DBLP:conf/iclr/ClarkLLM20}, to leverage contextual information. With \textsc{GloVe}, embeddings of the most frequent $50$ words in the training set are fixed during training while the remaining will be fine-tuned. The schema linking strategy is borrowed from RATSQL~\cite{wang-etal-2020-rat}, which is also our baseline system. During evaluation, we adopt beam search decoding with beam size $5$.

\paragraph{Hyper-parameters}
In the encoder, the GNN hidden size $d$ is set to $256$ for \textsc{GloVe} and $512$ for PLMs. The number of GNN layers $L$ is $8$. In the decoder, the dimension of hidden state, action embedding and node type embedding are set to $512$, $128$ and $128$ respectively. The recurrent dropout rate~\cite{DBLP:conf/nips/GalG16} is $0.2$ for decoder LSTM. The number of heads in multi-head attention is $8$ and the dropout rate of features is set to $0.2$ in both the encoder and decoder. Throughout the experiments, we use AdamW~\cite{DBLP:conf/iclr/LoshchilovH19} optimizer with linear warmup scheduler. The warmup ratio of total training steps is $0.1$. For \textsc{GloVe}, the learning rate is $5e\textrm{-}4$ and the weight decay coefficient is $1e\textrm{-}4$; For PLMs, we use smaller leaning rate $2e\textrm{-}5$~({\tt base}) or $1e\textrm{-}5$~({\tt large}), and larger weight decay rate $0.1$. The optimization of the PLM encoder is carried out more carefully with layer-wise learning rate decay coefficient $0.8$. Batch size is $20$ and the maximum gradient norm is $5$. The number of training epochs is $100$ for \textsc{Glove}, and $200$ for PLMs respectively.

\subsection{Main Results}
\begin{table}[htbp]
\centering {\small
    \begin{tabular}{c|cc}
    \toprule
    \textbf{Model} & \textbf{Dev} & \textbf{Test} \\
    \hline\hline
    \multicolumn{3}{c}{\textbf{Without PLM}} \\
    \hline
    GNN~\cite{bogin-etal-2019-representing} & 40.7  & 39.4  \\
    Global-GNN~\cite{bogin-etal-2019-global} & 52.7  & 47.4  \\
    EditSQL~\cite{zhang-etal-2019-editing} & 36.4  & 32.9  \\
    IRNet~\cite{guo-etal-2019-towards} & 53.2  & 46.7  \\
    RATSQL~\cite{wang-etal-2020-rat} & 62.7  & 57.2  \\
    \hline
    \textbf{LGESQL} & \textbf{67.6 } & \textbf{62.8} \\
    \hline\hline
    \multicolumn{3}{c}{\textbf{With PLM: \textsc{Bert}}} \\
    \hline
    IRNet~\cite{guo-etal-2019-towards} & 53.2  & 46.7  \\
    GAZP~\cite{zhong-etal-2020-grounded} & 59.1 & 53.3 \\
    EditSQL~\cite{zhang-etal-2019-editing} & 57.6  & 53.4  \\
    BRIDGE~\cite{lin-etal-2020-bridging} & 70.0  & 65.0  \\
    BRIDGE + Ensemble & 71.1 & 67.5 \\
    RATSQL~\cite{wang-etal-2020-rat} & 69.7  & 65.6  \\
    \hline
    \textbf{LGESQL} & \textbf{74.1} & \textbf{68.3} \\
    \hline\hline
    \multicolumn{3}{c}{\textbf{With Task Adaptive PLM}} \\
    \hline
    ShadowGNN~\cite{chen-etal-2021-shadowgnn} & 72.3 & 66.1 \\
    RATSQL+\textsc{Strug}~\cite{deng-etal-2021-structure} & 72.6 & 68.4 \\
    RATSQL+\textsc{Grappa}~\cite{DBLP:journals/corr/abs-2009-13845}  & 73.4  & 69.6 \\
    SmBoP~\cite{rubin-berant-2021-smbop} & 74.7 & 69.5 \\
    RATSQL+\textsc{Gap}~\cite{DBLP:journals/corr/abs-2012-10309} & 71.8  & 69.7  \\
    DT-Fixup SQL-SP~\cite{xu2021optimizing} & 75.0 & 70.9 \\ 
    \hline
    \textbf{LGESQL+\textsc{Electra}} & \textbf{75.1} & \textbf{72.0} \\
    \bottomrule
    \end{tabular}%
    \caption{Comparison to previous methods.}
  \label{tab:main}}
\end{table}%
The main results of the test set are provided in Table \ref{tab:main}. Our proposed line graph enhanced text-to-SQL~(LGESQL) model achieves state-of-the-art results in all configurations at the time of writing. With word vectors \textsc{GloVe}, the performance increases from $57.2\%$ to $62.8\%$, $5.6\%$ absolute improvements. With PLM {\tt bert-large-wwm}, LGESQL also surpasses all previous methods, including the ensemble model, and attains $68.3\%$ accuracy. Recently, more advanced approaches all leverage the benefits of larger PLMs, more task adaptive data~(text-table pairs), and tailored pre-training tasks. For example, \textsc{Gap}~\cite{DBLP:journals/corr/abs-2012-10309} designs some task adaptive self-supervised tasks such as column prediction and column recovery to better address the downstream joint encoding problem. We utilize {\tt electra-large} for its compatibility with our model and achieves $72.0\%$ accuracy.



Taking one step further, we compare more fine-grained performances of our model to the baseline system RATSQL~\cite{wang-etal-2020-rat} classified by the level of difficulty in Table \ref{tab:comp}. We observe that LGESQL surpasses RATSQL across all subdivisions in both the validation and test datasets regardless of the application of a PLM, especially at the \textbf{Medium} and \textbf{Extra Hard} levels. This validates the superiority of our model by exploiting the structural relations among edges in the line graph.
\begin{table}[htbp]
  \centering{\small
    \begin{tabular}{c|cccc|c}
    \toprule
    \multicolumn{1}{c}{\textbf{Split}} & \textbf{Easy} & \textbf{Medium} & \textbf{Hard} & \multicolumn{1}{c}{\textbf{Extra}} & \textbf{All} \\
    \hline
    \hline
    \multicolumn{6}{c}{\textbf{RATSQL}} \\\hline
    \textbf{Dev} & 80.4  & 63.9  & 55.7  & 40.6  & 62.7  \\
    \textbf{Test} & 74.8  & 60.7  & 53.6  & 31.5  & 57.2  \\
    \hline
    \multicolumn{6}{c}{\textbf{LGESQL}} \\\hline
    \textbf{Dev} & 86.3  & 69.5  & 61.5  & 41.0  & 67.6 \\
    \textbf{Test} & \textbf{80.9}  & \textbf{68.1} & \textbf{54.0} & \textbf{37.5} & \textbf{62.8} \\
    \hline
    \hline
    \multicolumn{6}{c}{\textbf{RATSQL+PLM:} {\tt bert-large-wwm}} \\
    \hline
    \textbf{Dev} & 86.4  & 73.6  & 62.1  & 42.9  & 69.7  \\
    \textbf{Test} & 83.0  & 71.3  & 58.3  & 38.4  & 65.6  \\
    \hline
    \multicolumn{6}{c}{\textbf{LGESQL+PLM:} {\tt bert-large-wwm}} \\\hline
    \textbf{Dev} & 91.5  & 76.7  & 66.7  & 48.8  & 74.1  \\
    \textbf{Test} &  \textbf{84.5} & \textbf{74.7} & \textbf{60.9} & \textbf{41.5} & \textbf{68.3} \\
    \bottomrule
    \end{tabular}%
  \caption{A detailed comparison to the reported results in the original paper RATSQL~\cite{wang-etal-2020-rat} according to the level of difficulty.}
  \label{tab:comp}}%
\end{table}%

\subsection{Ablation Studies}
In this section, we investigate the contribution of each design choice. We report the average accuracy on the validation dataset with $5$ random seeds.
\subsubsection{Different Components of LGESQL}
\begin{table}[htbp]
  \centering
    \begin{tabular}{c|c}
    \toprule
    \textbf{Technique} & \textbf{Dev Acc} \\
    \hline\hline
    \multicolumn{2}{c}{\bf Without Line Graph: RGATSQL} \\
    \hline
    w/ SE & 66.2 \\
    w/ MMC & 66.2 \\
    w/o NLC & 63.3 \\
    w/o GP & 65.5 \\
    \hline\hline
    \multicolumn{2}{c}{\bf With Line Graph: LGESQL} \\
    \hline
    w/ MSDE & 67.3  \\
    w/ MMC & 67.4  \\
    w/o NLC & 65.3  \\
    w/o GP & 66.2 \\
    \bottomrule
\end{tabular}
  \caption{Ablation study of different modules. SE: static embeddings; MMC: multi-head multi-view concatenation; MSDE: mixed static and dynamic embeddings; NLC: non-local relations; GP: graph pruning.}
  \label{tab:abl}
\end{table}%
RGATSQL is our baseline system where the line graph is not utilized. It can be viewed as a variant of RATSQL with our tailored grammar-based decoder. From Table \ref{tab:abl}, we can discover that: 1) if non-local relations or meta-paths are removed~(w/o NLC), the performance will decrease roughly by $2$ points in LGESQL, while $3$ points drop in RGATSQL. However, our LGESQL with merely local relations is still competitive. It consolidates our motivation that by exploiting the structure among edges, the line graph can capturing long-range relations to some extent. 2) graph pruning task contributes more in LGESQL~($+1.2\%$) than RGATSQL~($+0.7\%$) on account of the fact that local relations are more critical to structural inference. 3) Two strategies of combining local and non-local relations introduced in \cref{sec:mix}~(w/ MSDE or MMC) are both beneficial to the eventual performances of LGESQL~($2.0\%$ and $2.1\%$ gains, respectively). It corroborates the assumption that local and non-local relations should be treated with distinction. However, the performance remains unchanged in RGATSQL, when merging a different view of the graph~(w/ MMC) into multi-head attention. This may be caused by the over-smoothing problem of a complete graph.

\subsubsection{Pre-trained Language Models}
\begin{table}[htbp]
  \centering
    \begin{tabular}{c|c|c}
    \toprule
    \textbf{PLM} & \textbf{RGATSQL} & \textbf{LGESQL} \\
    \midrule
    {\tt bert-base} & 70.5  & 71.4  \\
    {\tt electra-base} & 72.8  & 73.4  \\
    \midrule
    {\tt bert-large} & 72.3  & 73.5  \\
    {\tt grappa-large} & 73.1  & 74.0  \\
    {\tt electra-large} & 74.8  & 75.1  \\
    \bottomrule
    \end{tabular}%
  \caption{Ablation study of different PLMs.}
  \label{tab:plm}%
\end{table}%
In this part, we analyze the effects of different pre-trained language models in Table \ref{tab:plm}. 
From the overall results, we can see that: 1) by involving the line graph into computation, LGESQL outperforms the baseline model RGATSQL with different PLMs, further demonstrating the effectiveness of explicitly modeling edge features. 2) {\tt large} series PLMs consistently perform better than {\tt base} models on account of their model capacity and generalization capability to unseen domains. 3) Task adaptive PLMs especially \textsc{Electra} are superior to vanilla \textsc{Bert} irrespective of the upper GNN architecture. We hypothesize the reason is that \textsc{Electra} is pre-trained with a tailored binary classification task, which aims to individually distinguish whether each input word is substituted given the context. Essentially, this self-supervised task is similar to our proposed graph pruning task, which focuses on enhancing the discriminative capability of the encoder.

\subsection{Case Studies}
\begin{figure}[htbp]
    \centering
    \includegraphics[width=0.48\textwidth]{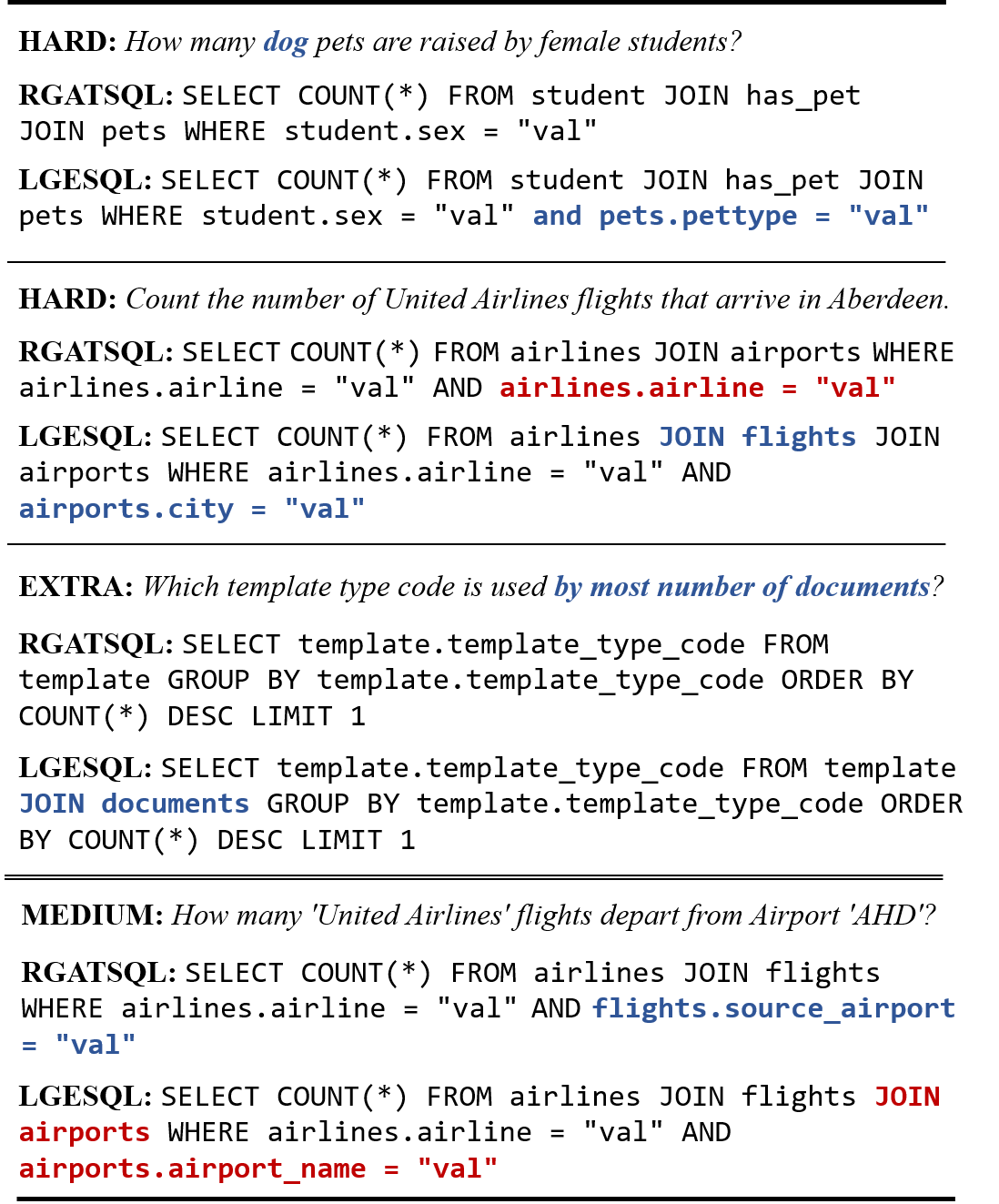}
    \caption{Case study: the first three cases are positive samples while the last one is negative. The input question is represented by its level of difficulty. {\tt FROM} conditions are omitted here for brevity and cell values in the SQL queries are replaced with placeholders ``val".}
    \label{fig:case}
\end{figure}
In Figure \ref{fig:case}, we compare the SQL queries generated by our LGESQL model with those created by the baseline model RGATSQL. We notice that LGESQL performs better than the baseline system, especially on examples that involve the {\tt JOIN} operation of multiple tables. For instance, in the second case where the connection of three tables are included, RGATSQL fails to identify the existence of table {\tt flights}. Thus, it is unable to predict the {\tt WHERE} condition about the destination city and does repeat work. In the third case, our LGESQL still successfully constructs a connected schema sub-graph by linking table ``template" to ``documents". Sadly, the RGATSQL model neglects the occurrence of ``documents" again. However, in the last case, our LGESQL is stupid to introduce an unnecessary table ``airports". It ignores the situation that table ``flights" has one column ``source\_airport" which already satisfies the requirement.

%% file: 5.related_work.tex
\section{Related Work}
\paragraph{Encoding Problem for Text-to-SQL}
To tackle the joint encoding problem of the question and database schema, \citet{xu2017sqlnet} proposes ``column attention" strategy to gather information from columns for each question word. TypeSQL~\cite{yu2018typesql} incorporates prior knowledge of column types and schema linking as additional input features. \citet{bogin-etal-2019-representing} and \citet{chen-etal-2021-shadowgnn} deal with the graph structure of database schema via GNN. EditSQL~\cite{zhang-etal-2019-editing} considers ``co-attention" between question words and database schema nodes similar to the common practice in text matching~\cite{chen-etal-2017-enhanced}. BRIDGE~\cite{lin-etal-2020-bridging} further leverages the database content to augment the column representation. The most advanced method RATSQL~\cite{wang-etal-2020-rat}, utilizes a complete relational graph attention neural network to handle various pre-defined relations. In this work, we further consider both local and non-local, dynamic and static edge features among different types of nodes with a line graph.

\paragraph{Heterogeneous Graph Neural Network}
Apart from the structural topology, a heterogeneous graph~\cite{shi2016survey} also contains multiple types of nodes and edges. To address the heterogeneity of node attributes, \citet{zhang2019heterogeneous} designs a type-based content encoder and \citet{fu2020magnn} utilizes a type-specific linear transformation. For edges, relational graph convolution network~(RGCN, \citealp{schlichtkrull2018modeling}) and relational graph attention network~(RGAT, \citealp{wang-etal-2020-relational}) have been proposed to parameterize different relations. HAN~\cite{wang2019heterogeneous} converts the original heterogeneous graph into multiple homogeneous graphs and applies a hierarchical attention mechanism to the meta-path-based sub-graphs. Similar ideas have been adopted in dialogue state tracking~\cite{chen2020schema,DBLP:journals/corr/abs-1905-11259}, dialogue policy learning~\cite{lc918-chen-coling18} and text matching~\cite{chen-etal-2020-neural-graph,lyu2021let} to handle heterogeneous inputs. In another branch, \citet{chen2017supervised}, \citet{8970828} and \citet{zhao-etal-2020-line} construct the line graph of the original graph and explicitly model the computation over edge features. In this work, we borrow the idea of a line graph and update both node and edge features via iteration over dual graphs.

%% file: 6.conclusion.tex
\section{Conclusion}
In this work, we utilize the line graph to update the edge features in the heterogeneous graph for the text-to-SQL task. Through the iteration over the structural connections in the line graph, local edges can incorporate multi-hop relational features and capture significant meta-paths. By further integrating non-local relations, the encoder can learn from multiple views and attend to remote nodes with shortcuts. In the future, we will investigate more useful meta-paths and explore more effective methods to deal with different meta-path-based neighbors.

%% file: 7.appendix.tex
\section{Local and Non-Local Relations}
\label{app:relations}
{\small
\begin{table*}[t]
  \centering
    \begin{tabular}{cccl}
    \hline
    Source $x$ & Target $y$ & Relation & Description \\
    \hline
    \textsc{Q} & \textsc{Q} & \textsc{Distance+1} & $y$ is the next word of $x$. \\
    \hline
    \textsc{C} & \textsc{C} & \textsc{ForeignKey} & $y$ is the foreign key of $x$. \\
    \hline
    \multirow{2}[2]{*}{\textsc{T}} & \multirow{2}[2]{*}{\textsc{C}} & \textsc{Has} & The column $y$ belongs to the table $x$.\\
      &   & \textsc{PrimaryKey} & The column $y$ is the primary key of the table $x$. \\
    \hline
    \multirow{3}[2]{*}{\textsc{Q}} & \multirow{3}[2]{*}{\textsc{T}} & \textsc{NoMatch} & No overlapping between $x$ and $y$.\\
      &   & \textsc{PartialMatch} & $x$ is part of $y$, but the entire question does not contain $y$. \\
      &   & \textsc{ExactMatch} & $x$ is part of $y$, and $y$ is a span of the entire question. \\
    \hline
    \multirow{4}[2]{*}{\textsc{Q}} & \multirow{4}[2]{*}{\textsc{C}} & \textsc{NoMatch} & No overlapping between $x$ and $y$. \\
      &   & \textsc{PartialMatch} & $x$ is part of $y$, but the entire question does not contain $y$. \\
      &   & \textsc{ExactMatch} & $x$ is part of $y$, and $y$ is a span of the entire question. \\
      &   & \textsc{ValueMatch} & $x$ is part of the candidate cell values of column $y$. \\
    \hline
    \end{tabular}%
  \caption{The checklist of all local relations used in our experiments. All relations above are asymmetric. For brevity, we only show one direction, and the opposite can be easily inferred. Q/T/C stands for \textsc{Question}/\textsc{Table}/\textsc{Column} node respectively.}
  \label{tab:local}%
\end{table*}%
}
In this work, meta-paths with length $1$ are local relations, and other meta-paths are non-local relations. Specifically, Table \ref{tab:local} provides the list of all local relations according to the types of source and target nodes. Notice that we preserve the \textsc{NoMatch} relation because there is no overlapping between the entire question and any schema item in some cases. This relaxation will dramatically increase the number of edges in the line graph. To resolve it, we remove edges in the line graph that the source and target nodes both represent relation types of \textsc{Match} series. In other words, we prevent information propagating between these bipartite connections during the iteration of the line graph.

The checklist in Table \ref{tab:local} is only a subset of all relations defined in RATSQL~\cite{wang-etal-2020-rat}. For the remaining relations, we treat them as non-local relations for a fair comparison to the baseline system RATSQL.

\section{Details of Text-to-SQL Decoder}
\label{app:decoder}
\subsection{ASDL Grammar}
\begin{figure*}[htbp]
    \centering
    \includegraphics[width=\textwidth]{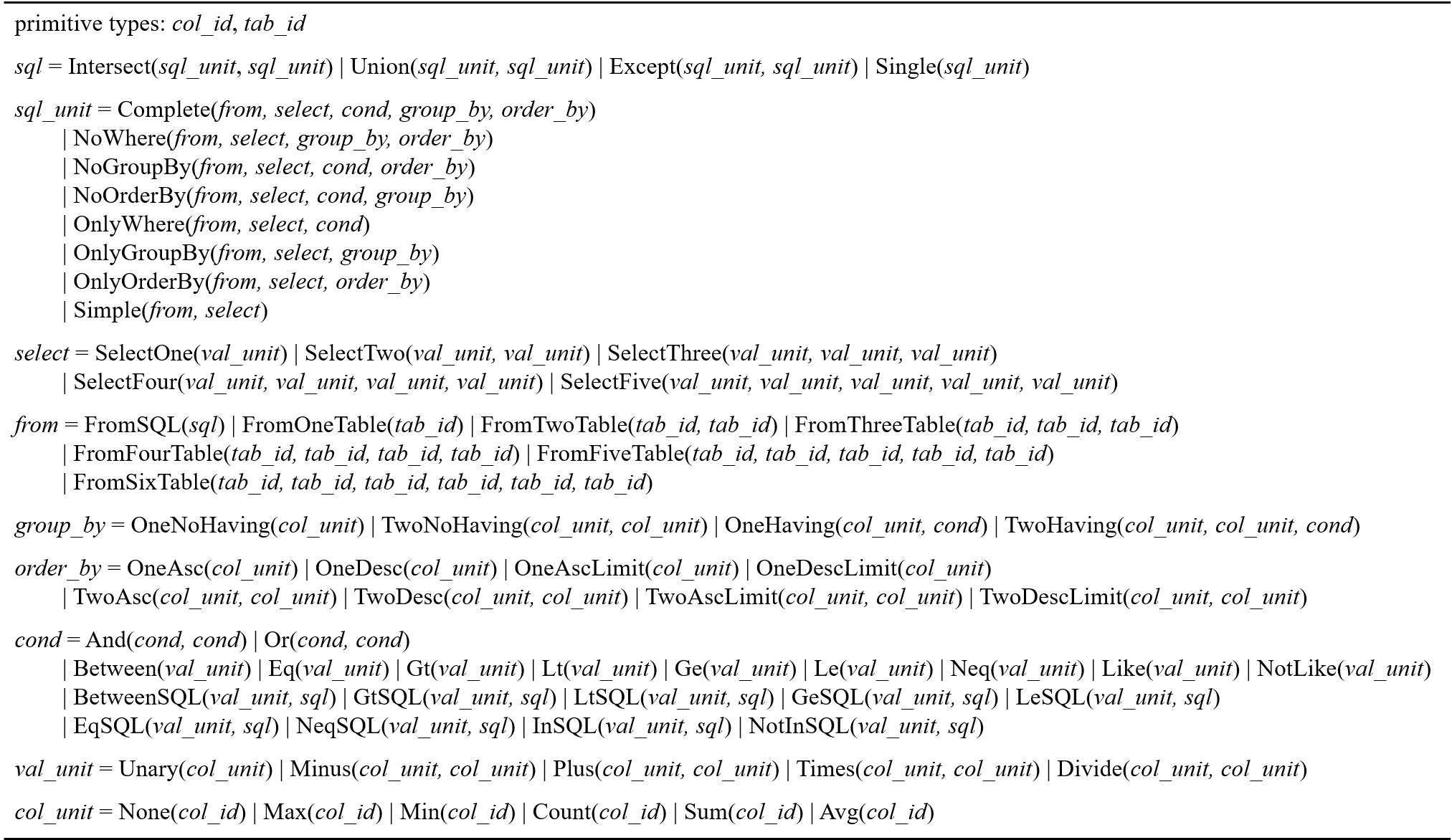}
    \caption{The ASDL grammar for SQL in our implementation.}
    \label{fig:grammar}
\end{figure*}
The complete grammar used to translate the SQL into a series of actions is provided in Figure \ref{fig:grammar}. Here are some criteria when we design the abstract syntax description language~(ASDL, \citealp{wang1997zephyr}) for the target SQL queries:
\begin{enumerate}
    \item Keep the length of the action sequence short to prevent the long-term forgetting problem in the auto-regressive decoder. To achieve this goal, we remove the optional operator ``?" defined in \citet{wang1997zephyr} and extend the number of constructors by enumeration. For example, we expand all solutions of type {\tt sql\_unit} according to the existence of different clauses.
    \item Hierarchically, group and re-use the same type in a top-down manner for parameter sharing. For example, we use the same type {\tt col\_unit} when choosing columns in different clauses and create the type {\tt val\_unit} such that both the SELECT clause and CONDITION clauses can refer to it.
    \item When generating a list of items of the same type, instead of emitting a special action \textsc{Reduce} as the symbol of termination~\cite{yin-neubig-2017-syntactic}, we enumerate all possible number of occurrences in the training set~(see the constructors for type {\tt select} and {\tt from} in Figure \ref{fig:grammar}). Then, we generate each item based on this quantitative limitation. Preliminary experimental results prove that thinking in advance is better than a lazy decision.
\end{enumerate}
Our grammar can cover $98.7\%$ and $98.2\%$ cases in the training and validation dataset, respectively.

\subsection{Decoder Architecture}
Given the encoded memory $\mathbf{X}=[\mathbf{X}_q;\mathbf{X}_t;\mathbf{X}_c]\in \mathbb{R}^{|V^n|\times d}$, where $|V^n|=|Q|+|T|+|C|$, the goal of a text-to-SQL decoder is to produce a sequence of actions which can construct the corresponding AST of the target SQL query. In our experiments, we utilize a single layer ordered neurons LSTM~(ON-LSTM,~\citealp{DBLP:conf/iclr/ShenTSC19}) as the auto-regressive decoder. Firstly, we initialize the decoder state $\mathbf{h}_0$ via attentive pooling over the memory $\mathbf{X}$.
\begin{align*}
a_i=&\text{softmax}_i\ \text{tanh}(\mathbf{x}_i\mathbf{W}_0)\mathbf{v}_0^{\mathrm{T}},\\
\tilde{\mathbf{h}}_0=&\sum_ia_i\mathbf{x}_i,\\
\mathbf{h}_0=&\text{tanh}(\tilde{\mathbf{h}}_0\mathbf{W}_1),
\end{align*}
where $\mathbf{v}_0$ is a trainable row vector and $\mathbf{W}_0,\mathbf{W}_1$ are parameter matrices. Then, in the structured ON-LSTM decoder, the hidden states at each timestep $j$ is updated as 
\begin{align*}
    \mathbf{m}_j,\mathbf{h}_j=\text{ON-LSTM}([\mathbf{a}_{j-1};\mathbf{a}_{p_j};&\mathbf{h}_{p_j};\mathbf{n}_{j}],\\
    &\mathbf{m}_{j-1},\mathbf{h}_{j-1}),
\end{align*}
where $\mathbf{m}_{j}$ is the cell state of the $j$-th timestep, $\mathbf{a}_{j-1}$ is the embedding of the previous action, $\mathbf{a}_{p_j}$ is the embedding of parent action, $\mathbf{h}_{p_t}$ is the embedding of parent hidden state, and $\mathbf{n}_{j}$ denotes the type embedding of the current frontier node~\footnote{The frontier node is the current non-terminal node in the partially generated AST to be expanded and we maintain an embedding for each node type.}. Given the current decoder state $\mathbf{h}_j$, we adopt multi-head attention~($8$ heads) mechanism to calculate the context vector $\tilde{\mathbf{h}}_j$ over $\mathbf{X}$. This context vector is concatenated with $\mathbf{h}_j$ and passed into a 2-layer MLP with tanh activation unit to obtain the attention vector $\mathbf{h}_{j}^{att}$. The dimension of $\mathbf{h}_{j}^{att}$ is $512$.

For \textsc{ApplyRule} action, the probability distribution is computed by a softmax classification layer:
\begin{multline*}
P(a_{j}=\textsc{ApplyRule}[R]|a_{<j},\mathbf{X})=\\
\text{softmax}_R(\mathbf{h}^{att}_j\mathbf{W}_{\text{R}}).
\end{multline*}

For \textsc{SelectTable} action, we directly copy the table $t_i$ from the encoded memory $\mathbf{X}_t$.
\begin{align*}
\zeta_{ji}^h=&\text{softmax}_{i} (\mathbf{h}^{att}_j\mathbf{W}^h_{tq})(\mathbf{x}_{t_i}\mathbf{W}^h_{tk})^{\mathrm{T}},\\
P(a_{j}=&\textsc{SelectTable}[t_i]|a_{<j},\mathbf{X})=\frac{1}{H}\sum_{h=1}^H\zeta^h_{ji}.
\end{align*}
To be consistent, we also apply the multi-head attention mechanism here with $H=8$ heads. The calculation of \textsc{SelectColumn} action is similar with different network parameters.

\section{Graph Pruning}
\label{app:gp}
Similar ideas have been proposed by \citet{bogin-etal-2019-global} and \citet{DBLP:journals/corr/abs-2009-13845}. Our proposed task differs from their methods in two aspects:

\paragraph{Prediction target}
\citet{DBLP:journals/corr/abs-2009-13845} devises several syntactic roles for schema items and performs multi-class classification instead of binary discrimination. Based on our assumption, the encoder is responsible for the discrimination capability while the decoder organizes different schema items and components into a complete semantic frame. Thus, we simplify the training target into binary labels.

\paragraph{Combination method}
\citet{bogin-etal-2019-global} utilizes another RGCN to calculate the relevance score for each schema item in Global-GNNSQL. This score is incorporated into the encoder RGCN as a soft input coefficient. Different from this cascaded method, graph pruning is employed in a multitasking manner. We have tried different approaches to combine this auxiliary module with the primary text-to-SQL model in our preliminary experiments, such as:

1) Similar to \citet{bogin-etal-2019-global}, we utilize a separate graph encoder to conduct graph pruning firstly, and use another refined graph encoder~(the same architecture, e.g., RGAT) to jointly encode the pruned schema graph and the question. These two encoders can share network parameters of only the embeddings or more upper GNN layers. If they share all $8$ layers, the entire encoder will degenerate from the pipelined mode into our multitasking fashion. Empirical results in Table \ref{tab:pipeline} demonstrate that when these two encoders share more layers, the performance of the text-to-SQL model is better.
\begin{table}[htbp]
  \centering
    \begin{tabular}{c|c|c}
    \toprule
    mode & \# layers shared & dev acc \\
    \midrule
    pipeline & 0 & 60.74 \\
\cmidrule{2-3}    $\Downarrow$ & 4 & 61.63 \\
\cmidrule{2-3}    multitasking & 8 & 62.53 \\
    \bottomrule
    \end{tabular}%
      \caption{Variation of performances when gradually increasing the number of layers shared between the pruning and the main encoders.}
  \label{tab:pipeline}%
\end{table}%

2) We can constrain the text-to-SQL decoder to only attend and retrieve schema items from the pruned encoded memory when calculating attention vectors and select columns or tables. In other words, the graph pruning module and the text-to-SQL decoder are connected in a cascaded way. Through pilot experiments, we observe the flagrant training-inference inconsistency problem. The text-to-SQL decoder is trained upon the golden schema items, but it depends on the predicted options from the graph pruning module during evaluation. Even if we endeavor various sampling-based methods~(such as random sampling, sampling from current module predictions, or sampling from neighboring nodes of the golden schema graph) to inject some noise during training, the performance is merely competitive to that with multitasking. Therefore, based on Occam's Razor Theorem, we only treat graph pruning as an auxiliary output module.